\def\year{2017}\relax
\documentclass[letterpaper]{article}
\usepackage{aaai17}
\usepackage{times}
\usepackage{helvet}
\usepackage{courier}
\usepackage{graphicx}
\usepackage{amsmath}
\usepackage{mathrsfs}
\usepackage{algorithm2e}
\usepackage{booktabs} 
\usepackage{CJKutf8}
\DeclareMathOperator*{\argmax}{arg\,max}

\begin{document} \begin{CJK*}{UTF8}{gbsn} 
\title{One Sentence One Model for Neural Machine Translation}
\author{Xiaoqing Li, Jiajun Zhang and Chengqing Zong\\
National Laboratory of Pattern Recognition, Institute of Automation\\ Chinese Academy of Sciences\\
\{xqli,jjzhang,cqzong\}@nlpr.ia.ac.cn\\
}
\maketitle
\begin{abstract}
Neural machine translation (NMT) becomes a new state-of-the-art and achieves promising translation results using a simple encoder-decoder neural network. This neural network is trained once on the parallel corpus and the fixed network is used to translate all the test sentences. We argue that the general fixed network cannot best fit the specific test sentences. In this paper, we propose the dynamic NMT which learns a general network as usual, and then fine-tunes the network for each test sentence. The fine-tune work is done on a small set of the bilingual training data that is obtained through similarity search according to the test sentence. Extensive experiments demonstrate that this method can significantly improve the translation performance, especially when highly similar sentences are available.

\end{abstract}

\section{Introduction}

\noindent Neural machine translation achieved great success recently \cite{blunsom-emnlp-13,sutskever-nips-14,bahdanau-iclr-15}. Thanks to the end-to-end training paradigm and the powerful modeling capacity of neural network, NMT can produce comparable or even better results than traditional statistical machine translation, only after a few years of development. However, it also raises some new problems, such as how to use open vocabulary and how to avoid repeating and missing translations. These problems have been addressed by various recent approaches \cite{luong-acl-15,jean-acl-15,tu-arxiv-16,mi-arxiv-16b}.

How to learn a good set of parameters is another challenge for nowadays deep neural networks. There has been some work in the field of NMT. Shen et al. \shortcite{shen2015minimum} propose to use task specific optimization function. Specially, they propose to directly optimize BLEU score instead of likelihood of the training data. Bengio et al. \shortcite{bengio2015scheduled} take search into consideration during training. In common practice, the decoder uses gold reference as history during training, but it has to use generated output as history during testing. To fix this discrepancy between training and testing, the authors propose to moderately replace gold reference with generated output during training. Wiseman and Rush \shortcite{wiseman2016sequence} take a similar approach and regard training as beam search optimization.

However, no matter how the network parameters are learnt, they are fixed after the training is finished in all current NMT practice. And the same model is applied to every testing sentence. A potential issue of this practice is that a neural network needs to be able to compress all translation knowledge into a fixed set of parameters, which is very hard in reality. So we propose to learn a specific model for each testing sentence by paying more attention to those related sentences. In particular, we propose a learning on-the-fly strategy for parameter fine-tuning. First, a general model is learnt from the whole training data. Then, for each testing sentence, we find some similar sentence pairs from the training data and use them to fine tune the parameters.

This procedure resembles how human do translation. Given a sentence, especially one we are not familiar, we always would like to search for some similar sentences and see how they are translated. Various translation knowledge can be learned from these examples, such as how to translate a lexicon or phrase in a specific context, and how to reorder the translation of different blocks according to some syntactic clues. Once our translation knowledge is refreshed, we can handle the sentence with much higher confidence. 

There are two key aspects for the method. One is how to define similarity and the other is how to find similar sentence pairs efficiently. For similarity measure, we tried string based similarity and hidden representation based similarity. Our approach has two additional steps compared with plain decoding: finding similar sentence pairs and fine tuning. To improve the efficiency, we used the technique of inverted index for fast retrieval. We also studied how the size of similar data influences the decoding time.

Experimental results show our approach can effectively improve the translation performance, especially when highly similar sentences are available. 

\begin{figure*}[t] 
\centering \includegraphics[width=0.6\textwidth]{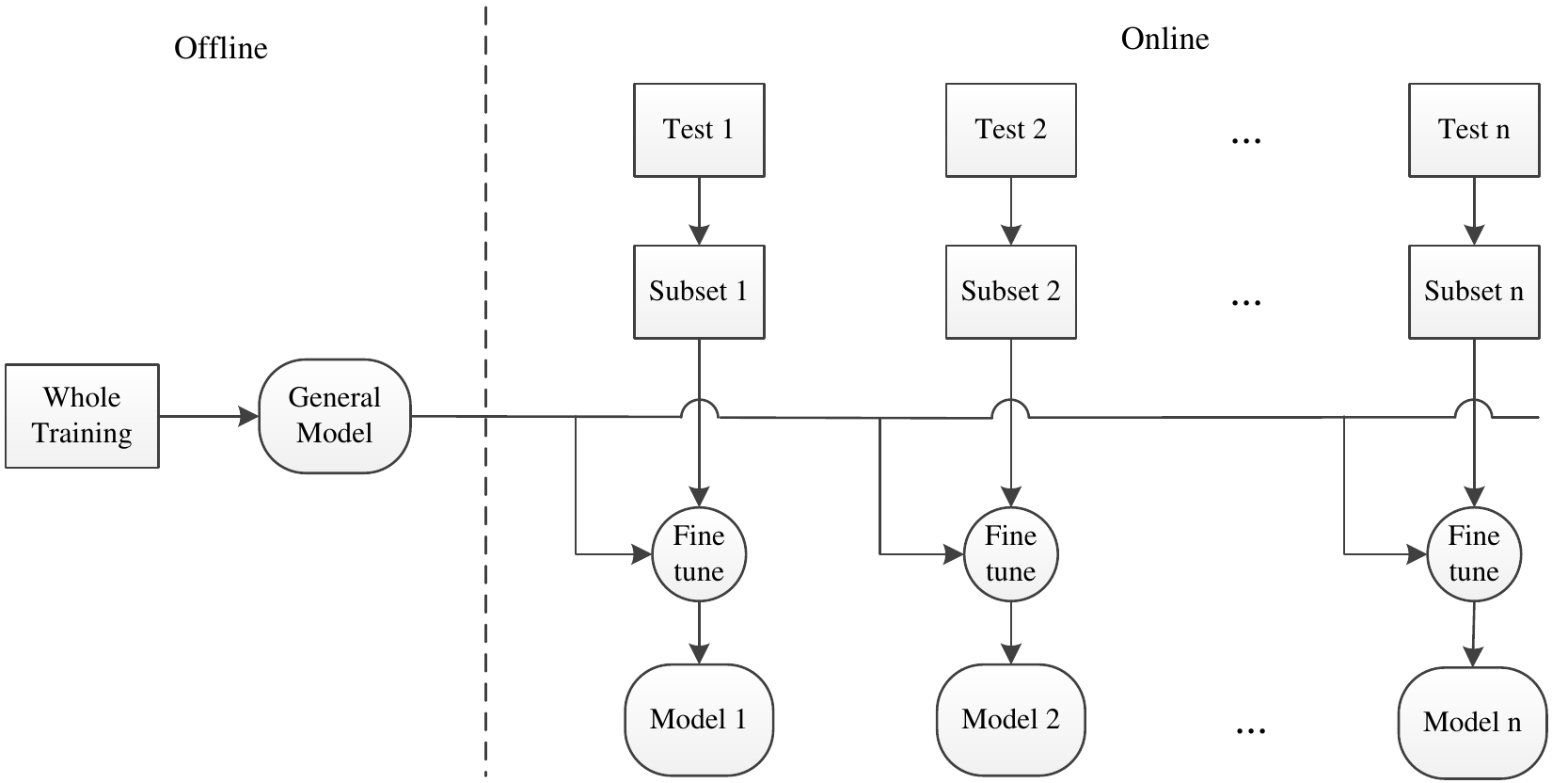}
\caption{System architecture for our method}
\end{figure*}

\section{Background}

In this section, we will briefly introduce the NMT system from Bahdanau et al. \shortcite{bahdanau-iclr-15}, which will be used later in the experiments. However, our approach is model independent and can be applied to other NMT systems.

Given a source sentence $s=(s_1,s_2,...s_m)$ and its translation $t=(t_1,t_2,...,t_n)$, NMT models the translation probability with a single neural network as follows,

\begin{equation}
p(t|s) = \prod_{i=1}^n p(t_i|t_{<i},s) 
\end{equation}

where the conditional probability is parameterized with the encoder-decoder framework. The encoder reads the source sentence and encodes it into a sequence of hidden states $h=h_1,h_2,...,h_m$ with bidirectional GRU.

\begin{equation}
    h_i = [\overrightarrow{h}_i;\overleftarrow{h}_i]
\end{equation}

\begin{equation}
\overrightarrow{h}_i = \overrightarrow{\phi}(\overrightarrow{h}_{i-1},x_i) 
\end{equation}

\begin{equation}
\overleftarrow{h}_i = \overleftarrow{\phi}(\overleftarrow{h}_{i+1},x_i) 
\end{equation}

where $x_i$ is the embedding of current word, and the recurrent activation functions $\overrightarrow{\phi}$ and $\overleftarrow{\phi}$ are gated recurrent units. 

The decoder consists of a recurrent neural network and an attention mechanism. The recurrent neural network computes a hidden state for each target position as follows,

\begin{equation}
    z_j = \phi(z_{j-1},y_{j-1},c_j)
\end{equation}
where $z_{j-1}$ is the previous hidden state, $y_{j-1}$ is the embedding of previous word and $c_j$ is the context vector obtained by the attention machenism,  which decides which source words to look at when predicting current target word.

\begin{equation}
    c_j = \sum_{i=1}^m \alpha_{i,j} h_i
\end{equation}

and the weight $\alpha_{i,j}$ is calculated as follows,

\begin{equation}
    \alpha_{i,j} = \frac{exp(e_{i,j})}{\sum_{k=1}^m exp(e_{k,j})}
\end{equation}
\begin{equation}
    e_{i,j} = f_{ATT}(z_{j-1},h_i)
\end{equation}

Then the probability of generating a specific target word $w$ will be computed by
\begin{equation}
    p(t_j = w|t_{<i},s) = softmax(z_j^\top y_w)
\end{equation}
where $y_w$ is the embedding of target word $w$.

\section{Tuning on-the-fly}

As illustrated in Figure 1, the learning strategy of our approach is simple. First, we learn a general model from the whole training corpus. Then, for each testing sentence, we extract a small subset from the training data, consisting of sentence pairs whose source sides are similar to the testing sentence. This subset is used to fine tune the general model and a specific model is obtained for the testing sentence. 

This procedure can be formulated as two stage optimization. The first stage is to to find a set of network parameters $\theta$ to maximize the log likelihood of the whole training data $D=\{(s^{(1)},t^{(1)}),(s^{(2)},t^{(2)}),...,(s^{(N)},t^{(N)})\}$.

\begin{align*}
\hat{\theta} &= \argmax_\theta \{\mathcal{L}(\theta)\} \\
&=  \argmax_\theta \{\log\prod_{k=1}^N P(t^{(k)}|s^{(k)};\theta)\} \\
&= \argmax_\theta \{\sum_{k=1}^N \sum_{i=1}^{|t^{(k)}|} \log P(t^{(k)}_i|s^{(k)},t^{(k)}_{<i};\theta)\}
\end{align*}

The second stage is to find a set of parameters in the neighbourhood of $\hat{\theta}$ to maximize the log likelihood of a subset of data similar to the testing sentence $x$.

\begin{align*}
\bar{\theta} =  \argmax_{\theta\in \mathcal{N}(\hat{\theta})} \{\log\prod_{s^{(k)}\sim s} P(t^{(k)}|s^{(k)};\theta)\} 
\end{align*}

In the following parts, we will discuss how to evaluate similarity between two sentences and how to quickly find similar sentences from training data.

\subsection{Similarity Measure}

There are many methods to evaluate the similarity between two sentences. In this paper, we consider three of them. The first is based on Levenshtein distance, which counts at least how many operations do we need to convert one sequence to another. The operations include insertion, deletion and substitution. Levenshtein distance reflects the surface similarity of two sentences, and it does not consider the meaning of the sentence.

\[
    sim_{LD}(s1,s2) = \frac{LD(s1,s2)}{max(len(s1),len(s2))}
\]

The second measure is based on average word embedding \cite{mikolov-nips-2013} of the sentence. Although this sentence representation is simple, it has been shown competitive to many complex sentence representations in many tasks.

%\begin{align*}
%    &sim_{vec}(s1,s2) \\
%    &= cos(\frac{\sum_{i=1}^{|s1|}vec(s1[i])}{len(s1)},\frac{\sum_{j=1}^{|s2|}vec(s2[j])}{len(s2)} )
%\end{align*}

\begin{align*}
    sim_{vec}(s1,s2) = cos(\frac{\sum_{i=1}^{|s1|}vec(s1[i])}{len(s1)},\frac{\sum_{j=1}^{|s2|}vec(s2[j])}{len(s2)} )
\end{align*}

The third measure is based on the hidden states of the encoder in NMT. Unlike word embedding, the hidden states of the encoder contains context information. What's more, the hidden states is learnt in the translation task. For this similarity measure, we need to run the encoder first with the general model learnt offline to get the representation of the testing sentence. This representation will be compared with the representation of training sentences, which need only to be calculated once in an offline manner.

\begin{align*}
    sim_{enc}(s1,s2)     = cos( \frac{\sum_{i=1}^{|s1|}h1[i]}{len(s1)},\frac{\sum_{j=1}^{|s2|}h2[j]}{len(s2)} )
\end{align*}

where $h1[i]$ and $h2[j]$ are the hidden states of the two sentences, which are calculated according to equations (2) - (4).

\subsection{Finding similar sentences efficiently}

The training corpus for neural machine translation usually contains millions of sentences. For a given testing sentence, comparing it with every training sentence will be too time consuming. So we propose to filter the training corpus first by only considering those which have common words with the testing sentence, and then compute similarity with the filtered set.

We use inverted index for fast retrieval. Each training sentence is given a unique index. And we maintain a word to indexes map, recording the sentence indexes where each word appears. For efficiency consideration, we ignore the most frequent words, which usually are function words and punctuations. Then for each word in a testing sentence, we find all sentences which contain the word. And the union of these sentences are used as the filtered set.

However, calculating Levenshtein distance between the testing sentence and each sentence in the filtered set is still not fast enough. So we propose to further reduce the set with a simpler similarity measure, i.e. dice coefficients. 
\[
    sim_{dice} = \frac{2|set(w\in s1)\cap set(w\in s2)|}{|set(w\in s1)|+|set(s\in s2)|}
\]

We first calculate the dice coefficients between the testing sentence and each sentence in the filtered set, then reduce the size of the set to a given threshold, e.g. 1000, by keeping the sentences with the highest dice coefficients. Finally, we will calculate Levenshtein distance for the reduced set.

For the other two similarity measures, calculating cosine similarity can be done efficiently with linear algebra library. So there is no need to further reduce the filtered set.

\subsection{Fine tuning}

The process of fine tuning is almost the same with offline training. The main difference is that the data size used for fine tuning is very small, usually containing only a few sentence pairs. So we need to be careful about overfitting. To this end, we go over the tuning data for only one pass. Learning rate is another factor need to be attended. Too large learning rate will cause overfitting, and too small learning rate will make it hard to learn translation knowledge from the tuning data. According to our pilot study, optimization methods with adaptive learning rate, such as Adadelta \cite{zeiler2012adadelta}, work as well as SGD with carefully tuned learning rate, so we adopt it in our experiments.

\section{Handle the case with low similarity}

We cannot always find very similar sentences to the testing sentence, especially when there is not enough in-domain training data.  In this case, we propose to find sentences to maximize phrase coverage. The phrase we mention here has the same meaning as the one in phrase-based machine translation, which denotes any consecutive word sequence. Our motivation is to select a subset of training data which can cover as many phrases in the testing sentence as possible.  The method to find the subset is shown in Algorithm 1.

\begin{algorithm}
 \KwIn{testing sentence $x$, training data $D$, phrase table $PT$}
 \KwOut{a subset of training data $D_x$ }
 $D_x \leftarrow \phi$;
 
\For{$i \leftarrow 1 $ \KwTo $ max\_phrase\_len$ }{
  \For{$j \leftarrow 1 $ \KwTo $ |x| - i$ }{
  check if $x[j:j+i]$ in $PT$\;
  \eIf{True}{
   \ForEach{$phrase \in phrase\_pairs$}{
   find a sentence pair containing $phrase$\;
   add the sentence pair to $D_x$\;
   }
   }{
   continue\;
  }
  }
  }
 \caption{Find a subset to maximize phrase coverage}
\end{algorithm}

The algorithm iterates over all possible phrases in the testing sentence and check if it is contained in the phrase table, which is extracted according to aligned bilingual corpus. The table contains a list of phrase pairs in the following form,
\begin{quote}
source $|||$ target $|||$ score1 score2 score3 score4
\end{quote}
The four scores for each phrase are direct phrase translation probability $\phi(t|s)$, inverse phrase translation probability $\phi(s|t)$, direct lexical weighting $lex(t|s)$, inverse lexical weighting $lex(s|t)$, which are used to evaluate the quality of the phrase pair from different angles. The direct phrase translation probability and lexical weighting are calculated as follows. The inverse ones are calculated similarly. 

\[
\phi(t|s) = \frac{Count(s,t)}{Count(s)}
\]
\[
lex(t|s) = \prod_{i=1}^{|t|}\frac{1}{|\{j|(i,j)\in a\}|}\sum_{\forall (i,j)\in a} p(e_i|f_j)
\]

A source phrase may corresponds to many (up to hundreds or thousands) target phrases, we filter them according to the average of the above four scores and keep those with the highest score. If a phrase in the testing sentence matches some source side in the phrase table, we will find a sentence pair in the training data which contains the source side and one of its high-score target side. Since there may be many sentence pairs containing such phrase pair, we choose one with the largest likelihood as follows, which means the sentence pair is simple and easy to learn.

\[
    (\hat{s},\hat{t}) = \argmax_{\textrm{phrase in } (s,t)} \prod_{i=1}^{|t|}P(t_i|t_{<i},s;\theta)
\]

The translation probability of each training sentence pair is calculated offline with the general network parameters. 

We don't use the phrase pairs as training data to fine-tune the network parameters. There are two reasons. First, context information is not available for choosing the proper phrase translation. Second, training on phrase pairs will harm the recurrent weights of the network, because they are not complete sentences\footnote{We also tried to fix the recurrent weights and tune the word embeddings only, it performs better than tuning all weights, but still worse than the approach of tuning on complete sentences.}.

\section{Experiments}

We evaluate our method on the Chinese to English translation task.
Translation quality is measured by the BLEU metric \cite{papineni-acl-02}.

\subsection{Datasets}

We conduct experiments on two datasets. One is on the United Nations Parallel Corpus\footnote{http://conferences.unite.un.org/UNCorpus}, which is composed of official records and other parliamentary documents of the United Nations. Since this data is from a narrow domain, we can easily find similar sentences for many testing sentences. The training data contains 1M sentence pairs extracted from the corpus, and the testing data contains 5 groups of sentence pairs, with 200 sentence pairs in each group. The most similar\footnote{The similarity is calculated based on Levenshtein distance.} sentence we can find for the sentences in each group falls into the similarity range of 0-0.2, 0.2-0.4, 0.4-0.6, 0.6-0.8 and 0.8-1.0, respectively. We also randomly selected 1,000 sentence pairs as development set. 

The training data of the other dataset is selected from LDC\footnote{https://www.ldc.upenn.edu/}, which contains about 1.2M sentence pairs, whose sources ranges from news, laws, hansard records, weblogs, spoken dialogues, etc. And we use NIST 03 as development set, and NIST 04 to 06 as testing set. In contrast to the UN data, we can hardly find very similar sentences to the testing one in this setting.

\begin{figure}[] \centering \includegraphics[width=0.40\textwidth]{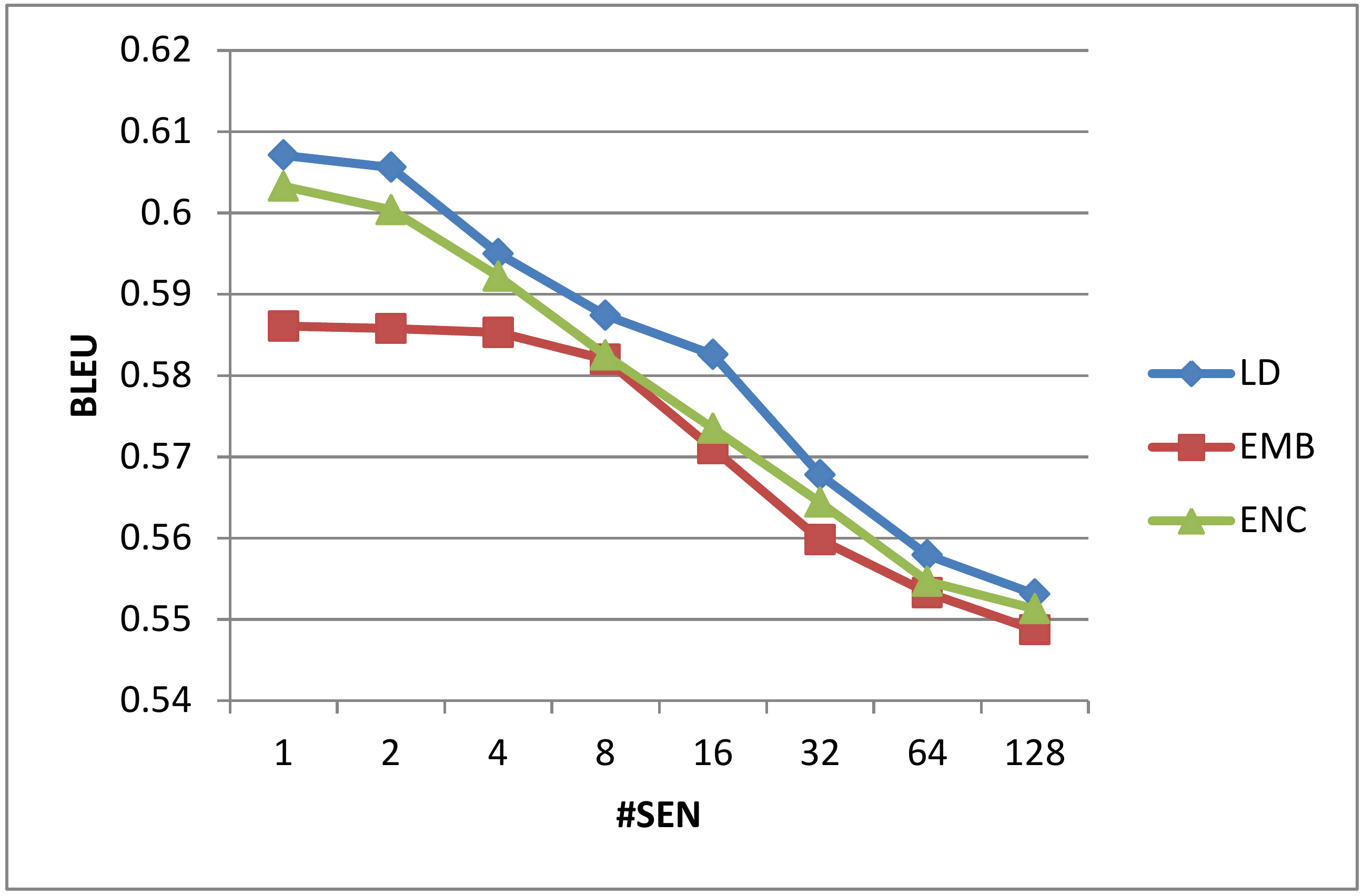}
\caption{Performance with different similarity measures when different number of similar sentences are used} 
\end{figure}

\begin{figure}[] \centering \includegraphics[width=0.40\textwidth]{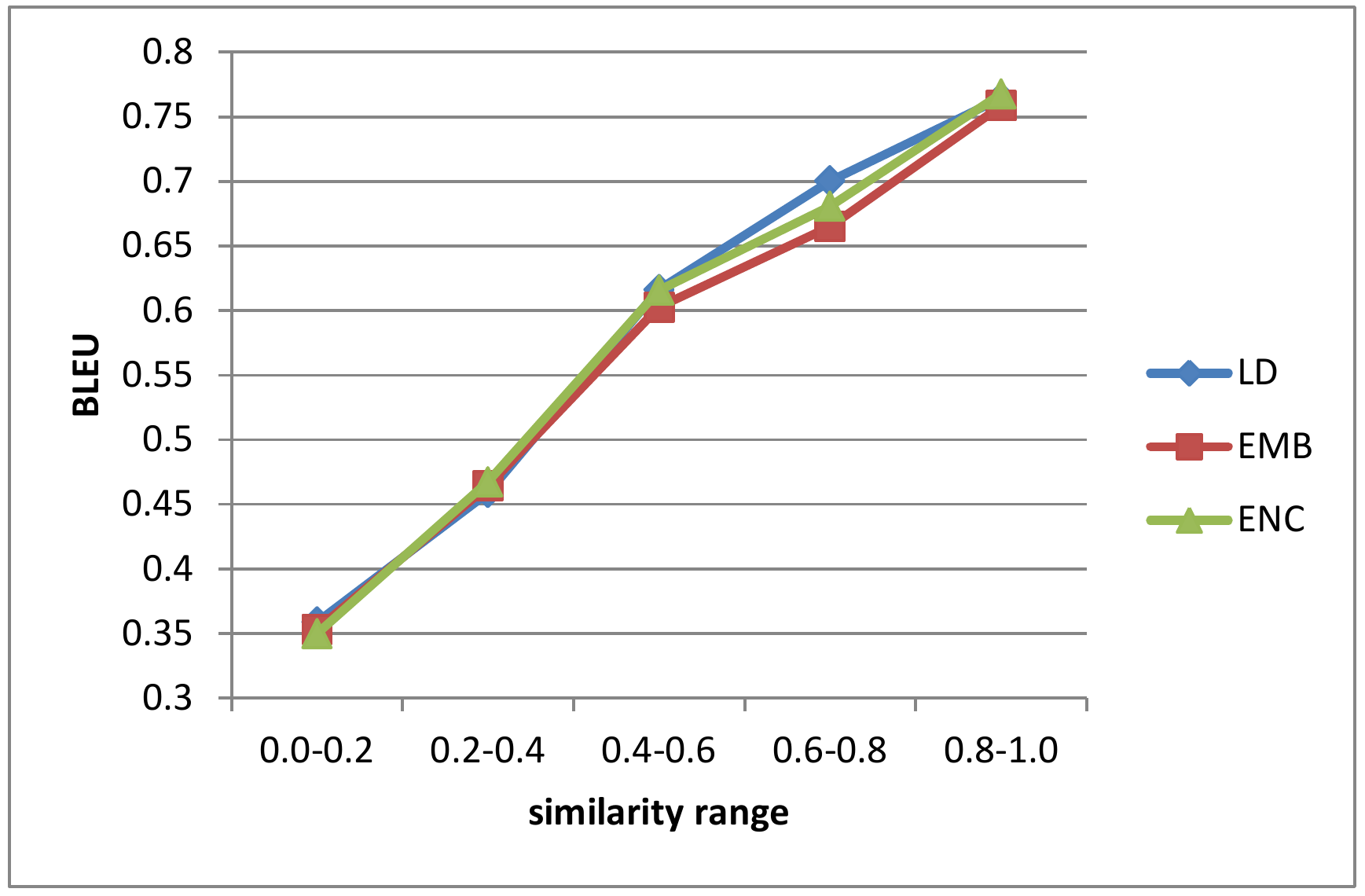}
\caption{Performance with different similarity measures on testing data with different similarity range} 
\end{figure}

\subsection{Experiment Setting}

The hyperparameters used in our network are described as follows. We limit both the source and target vocabulary to 30k in our experiments. The number of hidden units is 1,000 for both the encoder and decoder. And the embedding dimension is 500 for all source and target tokens. The network parameters are updated with the Adadelta algorithm for both training and fine tuning.

When finding similar sentences based on phrase coverage, we keep top two target phrase for each source phrase. And if a source phrase appears more than 1,000 times in the bilingual corpus, it will be discarded, because it's unnecessary to re-learn how to translate these common phrases.

\subsection{Experiments on UN Data}

We first conduct experiments of the UN corpus, studying which similarity measure is better, and how many similar sentences should be used. 

\subsubsection{Similarity Measure}

Figure 2 shows the performances of the three similarity measures when different number of similar sentences are used.  There are two observations according to this figure. First, the performance of the  similarity based on Levenshtein distance is always better than the other two, the similarity based on encoder states is slightly worse, and the similarity based on average word embedding is the worst. Since Levenshtein distance only cares string similarity, similar sentences found according to this measure will have more words in common with the testing sentence, thus more parameters related to the word embedding can be updated. And the encoder states takes context information into consideration when compared with averaged word embedding, so it has better performance.  

Second, the performance gap between different similarity measures become smaller when more similar sentences are used. This is due to the fact that there will be a larger overlapping in the sentences found by the three measures when more sentences are used.

To further check the difference between the three measures, we fix the number of sentences used for fine tuning as 4, and show the performance of the three measures on testing sentences in different similarity range in Figure 3. It can be seen from the figure that, when very un-similar (0-0.4) or very similar (0.8-1.0) sentences can be found for the testing sentence, the performances of the three measures have little difference. When the sentences in a relatively high range (0.4-0.8), especially in (0.6-0.8), can be found for the testing sentence, the performance of the Levenshtein distance based similarity is obviously better.

\begin{figure}[t] \centering \includegraphics[width=0.40\textwidth]{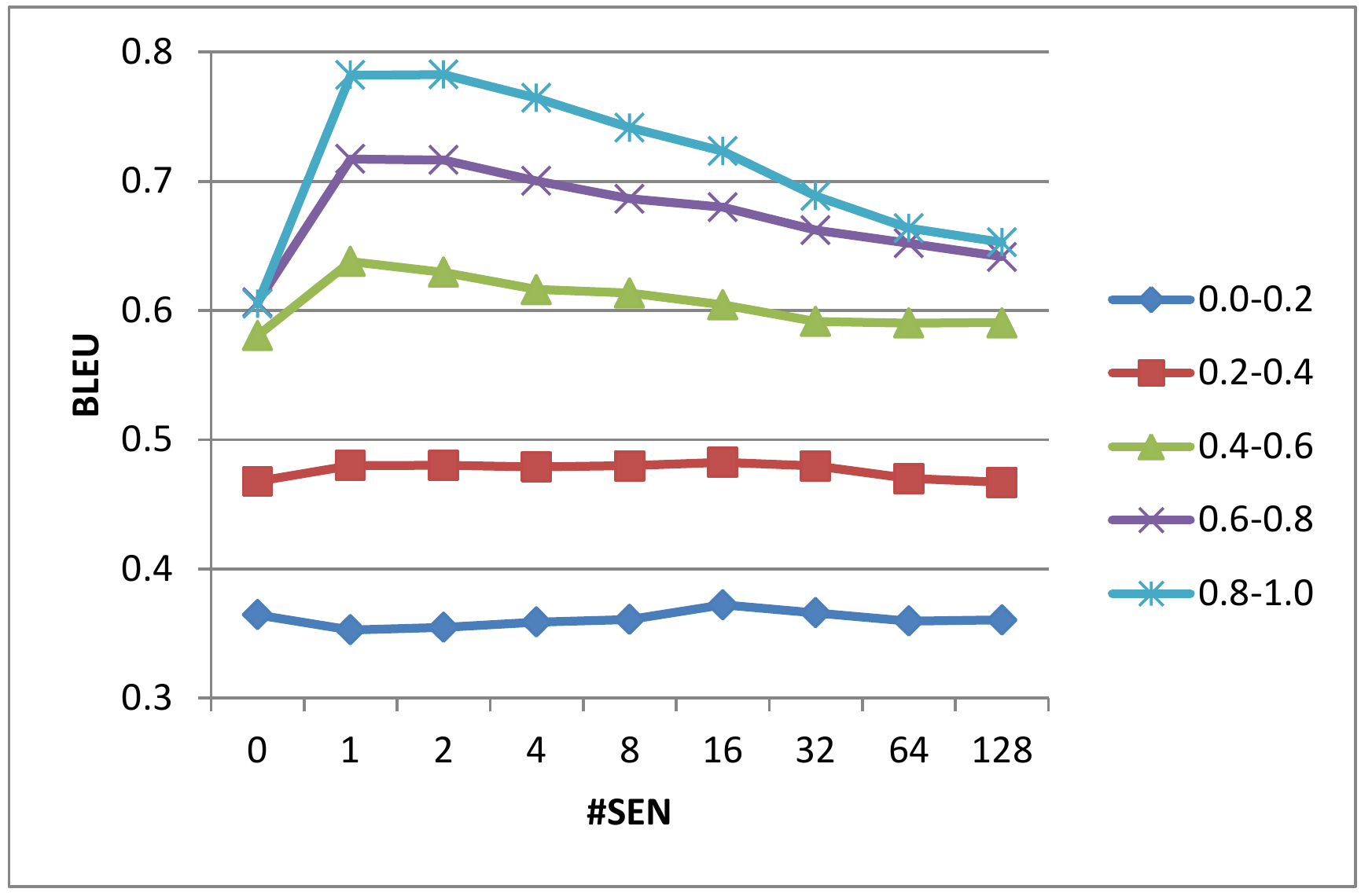}
\caption{How the size of similar data influence the performance of testing data with different similarity range} 
\end{figure}

\begin{table}[] 
    \centering 
    \begin{tabular}{llllll} 
        \toprule 
        System &  0-0.2 & 0.2-0.4 & 0.4-0.6 & 0.6-0.8 & 0.8-1\\ 
        \midrule 
      baseline & 36.45 & 46.78 & 58.06 & 60.64 & 60.52 \\
      fine-tune & 37.23 & 48.25 & 63.79 & 71.73 & 78.21 \\

        \bottomrule
      \end{tabular} 
      \caption{Best performance on each group of UN testing data} 
  \end{table}

\begin{figure}[t] \centering \includegraphics[width=0.40\textwidth]{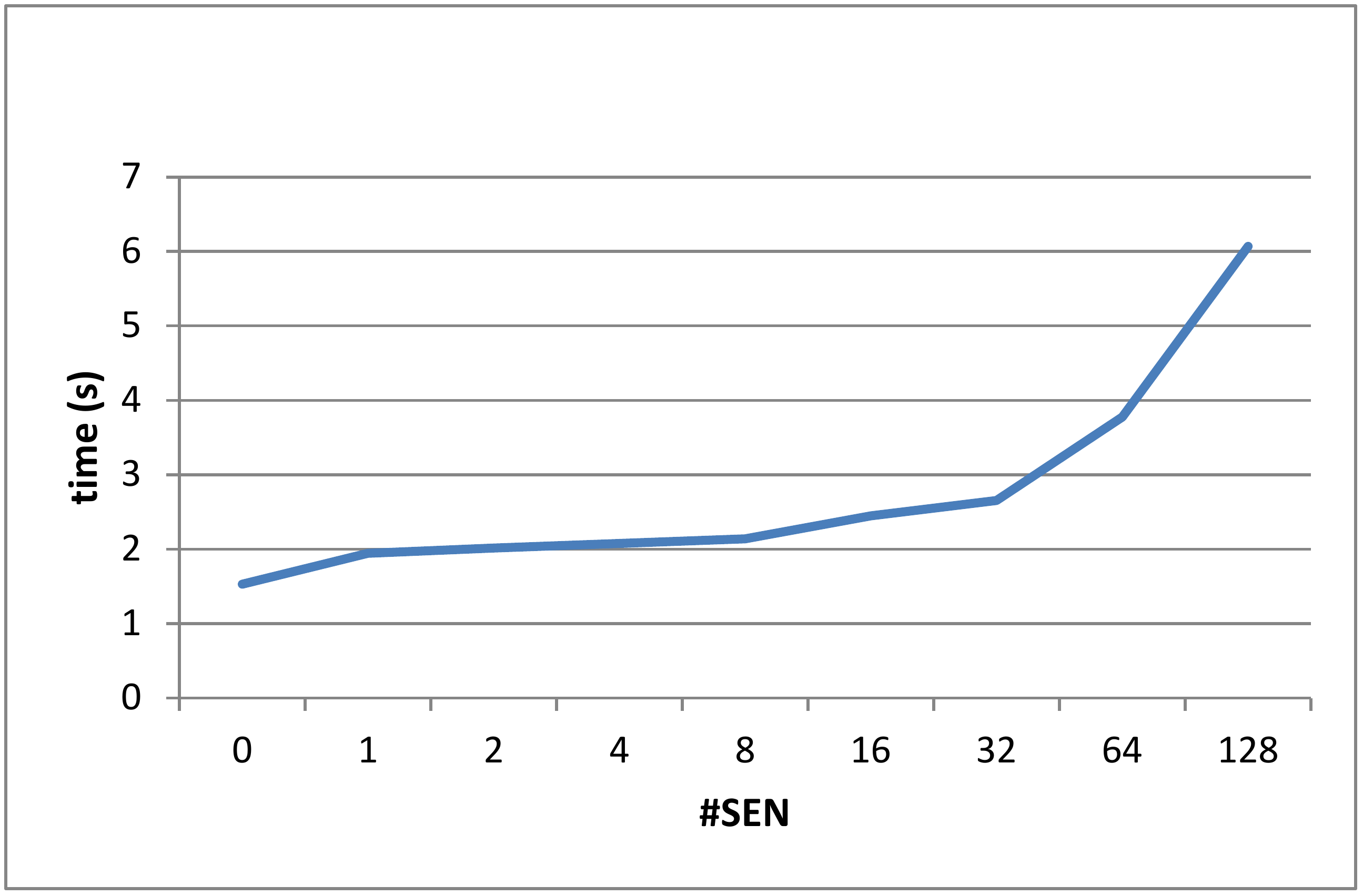}
\caption{How time cost increases while more sentences are used for fine-tuning} \end{figure}

  \begin{table}[] 
    \centering 
    \begin{tabular}{lllll} 
        \toprule 
        System &  04 & 05 & 06 & Avg.\\ 
        \midrule 
        baseline & 36.06      & 32.74 & 34.85 & 34.55 \\ 
        fine-tune & 37.43      & 34.01 & 35.77 & 35.74 \\ 
        fine-tune (phrase) & 38.04     & 34.41 & 35.41 & 35.99 \\ 
        \bottomrule
      \end{tabular} 
      \caption{Experimental results for LDC data} 
  \end{table}

\subsubsection{Data Size}

According to Figure 2, using only 1 similar sentence for fine-tuning performs best.  However, this figure only shows the overall performance on the whole testing data. If we dive into testing sentences with different similarity range, the trend will be different, as shown in Figure 4. We adopt the Levenshtein distance based similarity in this experiment. It can be seen from the figure that, if very similar sentences (0.4-1.0) can be found for the testing sentence, using only 1 similar sentence can greatly improve the performance, using more does not provide further help and may even degrade the performance. However, when the found sentences are not very similar (0-0.4), the improvement brought by fine-tuning is much smaller, and using more sentences, such as 16, is better than using one. Less sentences will lead to more severe overfitting, which will make the model remember how to reproduce the translations of the sentences. This is desirable when very similar sentences can be found, but it will produce negative effect otherwise.

The best performance we can get for each group of testing sentences are shown in Table 1. It can be seen from the table that more than 10 BLEU points can be gained when we can find very similar (0.6-1) sentences to the testing one. However, if we cannot find very similar sentences (0-0.4), only minor (around 1 BLEU point) improvement can be gained.

\begin{table*}[ht] 
    \centering 
    \begin{tabular}{lp{14cm}} 
        \toprule 
        input & 再次 要求 以色列 向 秘书长 提供 一切 便利 和 协助 , 以 执行 本 决议 \\
        reference & calls once again upon israel to render all facilities and assistance to the secretary - general in the implementation of the present resolution \\
        sim. & 再次 要求 以色列 向 秘书长 提供 一切 便利 以 执行 本 决议\\
        trans. of sim. & calls once more upon israel to render all facilities to the secretary - general in the implementation of the present resolution\\
        baseline & reiterates its request to the secretary - general to provide all facilities and assistance to the secretary - general for the implementation of the present resolution\\
        ours & calls once more upon israel to render all facilities and assistance to the secretary - general in the implementation of the present resolution\\
        \midrule 
        input & 经 讨论 商定 ， 去掉 方 括号 ， 保留 其中 的 内容 。\\
        reference & after discussion it was agreed to delete the square brackets and retain the contents therein . \\
        sim. & 工作组 商定 ， 去掉 该 款 的 方 括号 。 \\
        trans. of sim. & the working group agreed to remove square brackets from this paragraph .  \\
        baseline & after discussion , it was agreed that the removal of the content would be deleted . \\
        ours & after discussion , it was agreed to remove square brackets and retain the contents of it .\\
        \bottomrule
      \end{tabular} 
      \caption{Translation examples of our method} 
  \end{table*}

\subsubsection{Influence on Efficiency}

The influence of data size on efficiency is shown in Figure 5. The time cost in the figure only includes fine-tuning time and decoding time. The retrieval time, i.e., time of finding similar sentences, is not shown because it is relatively small compared to the other two. Retrieval with edit distance measure is the slowest one. But it is still less than 1/3 of the decoding time. We can see from the figure if less or equal than 32 sentences are used for fine-tuning for each testing sentence, the time cost is controlled within two times of the baseline. If we use 128 sentences, the time cost increases to 4 times.

\subsection{Experiments on LDC Data}

In this experiment, we can only find similar sentences in the range of 0-0.4 for more than 90\% of the testing sentences. And according to our study on the development set, the number of sentences used for fine-tuning needs to be increased to 128 to get the best performance when the similarity is low. We think the reason is due to the diversity of the training data. Sentences in the low similarity range may have totally different topics and styles with the testing one. In order to avoid the influence of these unwanted data, more sentences need to be used.

The performances on the testing data are shown in Table 2. They are obtained with the following setting, if very similar sentences (0.4-1) can be found, we use only 1 sentence for fine-tuning, otherwise we use 128 sentences. On average, 1.2 BLEU points can be gained on the three testing sets, which is consistent with the experimental results on the UN dataset when very similar sentences cannot be found.

The performances of finding similar sentences based on phrase coverage are also shown in the table. The average improvement is 1.45 BLEU points,  slightly better than the approach of finding similar sentences directly. And the average sentence number used for fine-tuning is 31, much less than 128. So the time cost is almost halved (see Figure 5).

\subsection{Result Analysis}

We show two examples in Table 3. The above one is the case where highly similar sentence can be found to the testing sentence. After fine-tuning, the model remembers how to generate the translation for the similar sentence. Based on the backbone, it can produce a correct translation for the testing sentence with a minor modification. In the lower example, we can only find a not so similar sentence to the testing one. However, the sentence pair found in the example can remind the model how to translate the phrase ``方括号'', whose translation is missing in the baseline system. 

\section{Related Work}

Neural machine translation has a short history of only a few years.
Kalchbrenner  and  Blunsom \shortcite{blunsom-emnlp-13} and Cho et al. \shortcite{cho-emnlp-14} first propose to use the
encoder-decoder architecture to do sequence to sequence mapping.  At the same time, Sutskever et al. \shortcite{sutskever-nips-14} apply it in end-to-end machine translation. 
Bahdanau et al. \shortcite{bahdanau-iclr-15} propose the attention mechanism to dynamically
attend to different source words when generating different target words, which becomes the default component of current NMT systems.

Recent advances in NMT include fixing defects of the model, such as inability to use large vocabulary \cite{luong-acl-15,jean-acl-15}, unawareness of coverage  \cite{tu-arxiv-16,mi-arxiv-16b} etc, making use of mono-lingual data \cite{cheng2016semi,sennrich2015improving}, extending to multi-lingual\cite{dong2015multi,zoph2016multi} and multi-modal \cite{hitschler2016multimodal} scenarios.

In statistical machine translation, there are some work making use of similar sentences by means of translation memory \cite{koehn2010convergence,ma2011consistent,wang2013integrating,li2014discriminative}. However, they need carefully designed features and only show improvement when similarity level is high. In comparison, our method don't need any modification to the model, and it can bring improvement in all similarity level.

Finding similar sentences with inverted index is fast enough in our experiments. If the training data is much larger than ours, locality sensitive hash such as MinHash \cite{broder1997resemblance} may be a better choice.

\section{ Conclusion}

In this paper, we propose to learn a specific model for each testing sentence. This is accomplished by two-stage training. An general model is learnt offline on the whole bilingual training corpus. During testing, a small batch of similar sentences are extract to fine-tune the network parameters on-the-fly. Experimental results demonstrate the effectiveness of this approach. When highly similar sentences are available, the improvement can exceed 10 BLEU points. Since our method is model independent, it can also be applied to other tasks beyond machine translation.

\bibliography{aaai}
\bibliographystyle{aaai} 
\end{CJK*}
\end{document}